\documentclass{llncs}

\newcommand{\repeatthanks}{\textsuperscript{\thefootnote}}

\usepackage{amsfonts}
\usepackage{graphicx}
\usepackage{amsmath}
\usepackage{multirow}

\usepackage{color}
\usepackage[misc]{ifsym}

\begin{document}

\title{Deep Learning Based Instance Segmentation in 3D Biomedical Images Using Weak Annotation}
\author{Zhuo Zhao$^{1,}\thanks{These authors contributed equally to
this work.}$, Lin Yang$^{1,}$\repeatthanks, Hao Zheng$^1$, Ian H. Guldner$^2$, Siyuan Zhang$^2$, and Danny Z. Chen$^1$}
\institute{$^1$ Department of Computer Science and Engineering,\\ University of Notre Dame, Notre Dame, IN 46556, USA\\
$^2$ Department of Biological Sciences, Harper Cancer Research Institute,\\ University of Notre Dame, Notre Dame, IN 46556, USA}

\maketitle

\begin{abstract}

Instance segmentation in 3D images is a fundamental task in biomedical image analysis. While deep learning models often work well for 2D instance segmentation, 3D instance segmentation still faces critical challenges, such as insufficient training data due to various annotation difficulties in 3D biomedical images. Common 3D annotation methods (e.g., full voxel annotation) incur high workloads and costs for labeling enough instances for training deep learning 3D instance segmentation models. In this paper, we propose a new weak annotation approach for training a fast deep learning 3D instance segmentation model without using full voxel mask annotation. Our approach needs only 3D bounding boxes for all instances and full voxel annotation for a small fraction of the instances, and uses a novel two-stage 3D instance segmentation model utilizing these two kinds of annotation, respectively. We evaluate our approach on several biomedical image datasets, and the experimental results show that (1) with full annotated boxes and a small amount of masks, our approach can achieve similar performance as the best known methods using full annotation, and (2) with similar annotation time, our approach outperforms the best known methods that use full annotation.

\end{abstract}

\section{Introduction}



3D instance segmentation seeks to segment all instances of the objects of interest (RoI) 
in 3D images.
This is a fundamental task in computer vision and biomedical image analysis. 
Recent successes at acquiring 3D biomedical image data \cite{ulman2017objective}
put even higher demand on 3D instance segmentation. However, annotation of 3D biomedical images to produce sufficient training data for deep learning models is often highly expensive and time-consuming, because only experts can annotate biomedical images well and no direct annotation technique is yet available for 3D biomedical images.
Further, although many 2D weakly supervised methods \cite{hu2017learning,khoreva2017simple,yang2017suggestive,zhang2017deep} were developed to reduce annotation efforts, they are not directly applicable to 3D images.
A common way to label 3D biomedical images is full annotation (i.e., all voxels of all RoI instances are annotated). This may work for voxel-level 3D segmentation networks \cite{chen2016voxresnet}, but 
instance segmentation demands much higher workload to annotate a sufficient (large) amount of instances for model training. Hence, using full annotation for 3D instance segmentation is impractical and annotation difficulties are a major obstacle impeding the development of deep learning models for 3D instance segmentation.

Recent 2D instance detection and segmentation methods achieved good performance \cite{girshick2015fast,he2017mask,liu2016ssd,ren2015faster}. However, in addition to the above annotation difficulties for 3D instance segmentation, extending such 2D approaches to 3D directly faces considerable challenges (e.g., GPU memory limit). In \cite{yang20163d}, 2D pixel segmentation results were stacked as 3D voxel segmentation results, and an algorithm (of high complexity) was then applied to the voxel segmentation results to conduct 3D instance segmentation. Although 2D annotation and the 2D model \cite{yang20163d} did not suffer GPU memory issues as much, without taking advantage of 3D context information, the stacked voxel segmentation results were not very accurate, and due to high algorithm complexity \cite{yang20163d}, processing a dense 3D stack took hours. 

\begin{figure}[t]
	\centering
	\includegraphics[width=10cm]{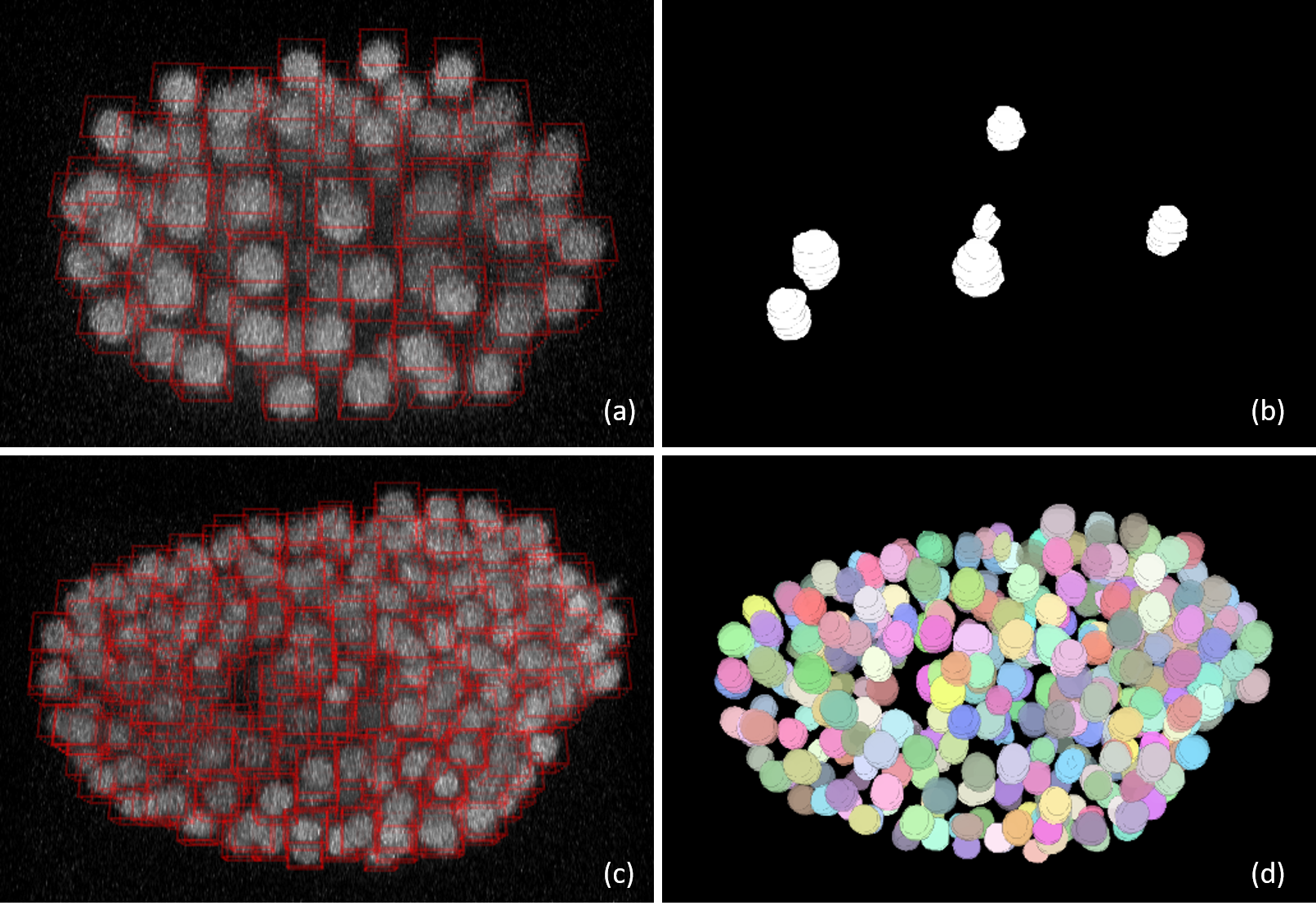}
	\caption[ ]{A training image example and test results for C.elegans developing embryos \cite{ulman2017objective}. (a) Each object instance is labeled by a 3D bounding box; (b) a small fraction of instances are labeled with voxel mask annotation; (c) in stage one, our model uses full box annotation to detect all instances; (d) in stage two, it uses full voxel mask annotation for a small fraction of the instances to segment each detected instance.}
	\label{fig:gt_test}
\end{figure}

To train a fast 3D instance segmentation model without high 3D annotation effort, in this paper, we present an end-to-end deep learning 3D instance segmentation model utilizing weak annotation. Our model needs only 3D bounding boxes for all instances and full voxel annotation for a small amount of instances (Fig.~\ref{fig:gt_test}(a)-(b)). The model has two stages. In the first stage, the model detects all instances utilizing 3D bounding box annotation; in the second stage, the model segments all detected instances utilizing full voxel annotation for a small amount of instances (Fig.~\ref{fig:gt_test}(c)-(d)). We adopt the design of VoxRes block \cite{chen2016voxresnet} to allow information propagating directly in both the forward and backward directions.

We evaluate our 3D instance segmentation approach on several datasets, and the experimental results show that (1) with full annotated boxes and a small amount of masks, our approach can achieve similar performance as the best known methods using full annotation, and (2) with similar annotation time, our approach outperforms the best known methods that use full annotation.

\section{Method}

Our proposed method consists of two major components: (1) a 3D object detector utilizing 3D bounding box annotation for all instances to predict 3D bounding boxes along with the probabilities of the boxes containing instances; (2) a 3D voxel segmentation model utilizing full voxel annotation for a small amount of instances to segment all instances of all objects of interest (RoI).

\subsection{3D object detector using 3D bounding box annotation}

We first briefly review 2D region proposal networks (RPN) for object detection, and then present our 3D object detector utilizing 3D bounding box annotation.

\noindent
\textbf{Region proposal networks (RPN)}. For object detection tasks, there are often two major steps: generating region proposals and classifying the proposals into different classes. 
Faster-RCNN \cite{girshick2015fast} proposes an FCN based RPN to generate RoIs from convolutional feature maps directly, and then uses a classifier to classify the generated RoIs into different classes. Different from the previous FCN models for predicting the probability for each pixel, RPN predicts the probabilities for the anchor boxes centering at each feature point (of an instance) 
and the box regression offset. A multi-task loss for RPN is defined as:
\begin{equation}
    L({p_i},{t_i}) = \frac{1}{N}\sum_{i}L_{cls}(p_i, p_i^*)+\lambda\frac{1}{N}\sum_{i}p_i^*L_{reg}(t_i, t_i^*)
\end{equation}
where $i$ is the index of an anchor box, $p_i$ is the predicted probability of anchor $i$, $p_i^*$ is a ground-truth label (0 for negative, 1 for positive), $t_i$ is a 4D vector presenting the shape of the predicted box (two for the box center position and two for the box size), $t_i^*$ presents the ground-truth box associated with a positive anchor, 
$L_{cls}$ is the log loss over two classes (object and background) for the objectness error, and $L_{reg}$ is the smooth $L_1$ loss \cite{girshick2015fast} on $t_i$ and $t_i^*$ for box position error. $L_{cls}$ and $L_{reg}$ are normalized by the number of the anchor boxes $N$.

\noindent
\textbf{3D object detector using 3D bounding box annotation}.
Given the fact that labeling a 3D bounding box for each instance is much cheaper than labeling all voxels of the instance, we propose a 3D object detector utilizing 3D bounding box annotation to detect all the instances in a 3D stack.

\begin{figure}[t]
	\centering
	\includegraphics[width=10cm]{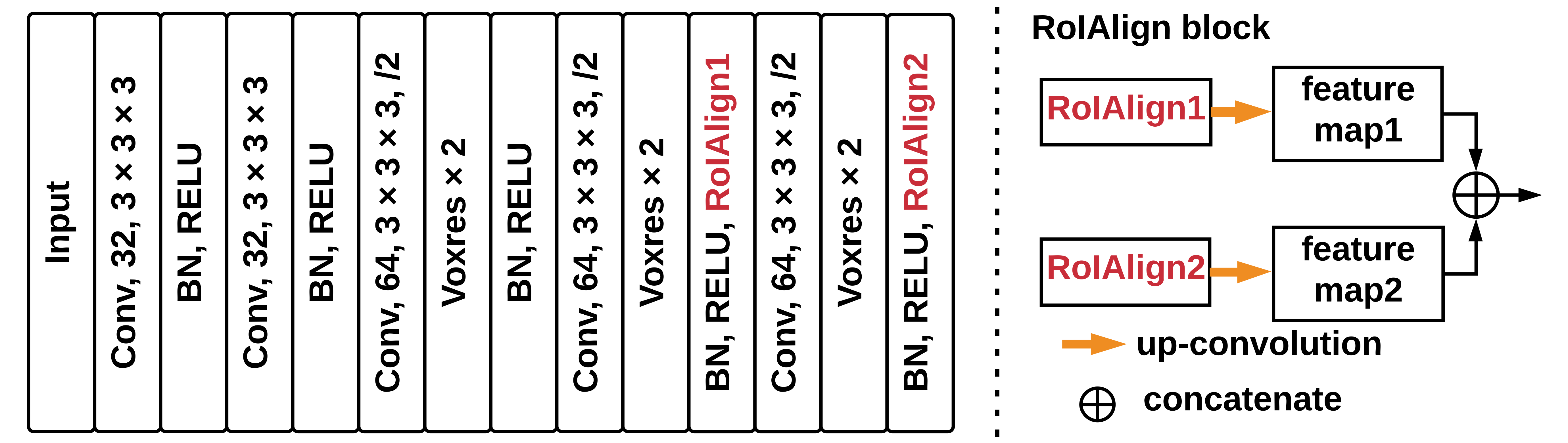}
	\caption[ ]{The backbone and RoIAlign block of the model. RoIAlign is applied to two layers of the backbone. The feature maps after up-convolution are concatenated as the final feature maps.}
	\label{backbone}
\end{figure}

Fig.~\ref{backbone} shows our FCN based backbone. We adopt the VoxRes block design to allow information propagating in both forward and backward directions directly. 

After extracting the feature maps from the backbone, we evaluate a fix number of anchor boxes of different sizes at each location of the feature maps.
We first match the anchor boxes to the ground-truth boxes by calculating the maximal Intersection of Union (IoU) between an anchor box and all ground-truth boxes. 
The anchor boxes having the maximal IoU with some ground-truth boxes are treated as positives, and all the anchor boxes having an IoU over a threshold (0.4 in our experiments) with any ground-truth boxes are also treated as positives. All other boxes are treated as negatives. 

\begin{figure}[t]
	\centering
	\includegraphics[width=10cm]{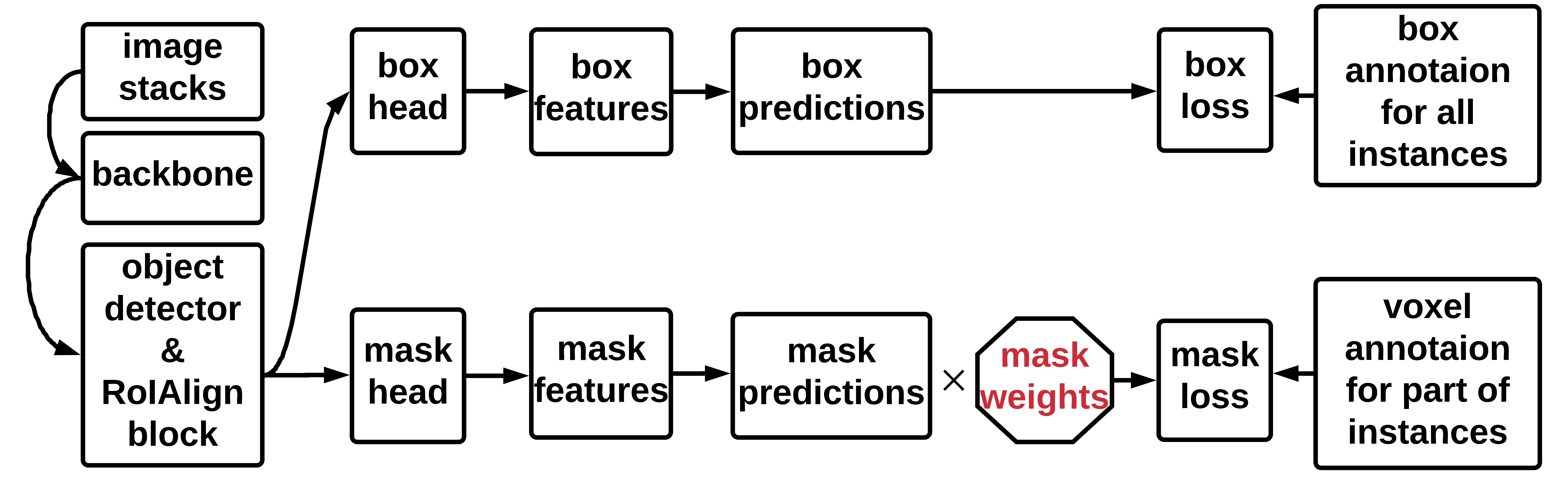}
	\caption[ ]{Illustrating the flow of our method. All boxes contribute to the object detector; only the instances with voxel annotation contribute to the voxel segmentation model.}
	\label{flow}
\end{figure}

A group of convolutional filters is used to predict the shape offset and score of containing an object for all anchor boxes. For a feature layer of size $m\times n\times k$ with $p$ channels, $cr$ filters ($c$ is the number of classes, $r$ is the number of anchor boxes for each feature point) of size $3\times 3\times 3\times p$ are used to predict the instance classes, and $6r$ filters of size $3\times 3\times 3\times p$ are used to predict the box regression offset of all anchor boxes. Hence, the output size for the score of the instance class is $m\times n\times k\times cr$, and the output size for the  box regression offset is $m\times n\times k\times 6r$.
The regression value $(t_z, t_x$, $t_y, t_d, t_h, t_w)$ of the predicted 3D box and the value $(t_z^*, t_x^*, t_y^*, t_d^*, t_h^*, t_w^*)$ of the ground-truth box  are computed as:
\begin{equation}
    \begin{aligned}
     t_z = (z-z_a)/d_a, t_x = (y-y_a)/h_a, t_y = (x-x_a)/w_a,\\
     t_d = \log(d/d_a), t_h = \log(h/h_a), t_w = \log(w/w_a),\\
     t_z^* = (z^*-z_a)/d_a, t_x^* = (y^*-y_a)/h_a, t_y^* = (x^*-x_a)/w_a,\\
     t_d^* = \log(d^*/d_a), t_h^* = \log(h^*/h_a), t_w^* = \log(w^*/w_a),
    \end{aligned}
\end{equation}
where $z$, $x$, and $y$ denote the box center coordinates, $d$, $h$, and $w$ denote the box size, and $z$, $z_a$, and $z^*$ are for the predicted box, anchor box, and ground-truth box, respectively (likewise for $x$, $y$, $d$, $w$, and $h$). All anchor boxes are regressed to nearby ground-truth boxes to calculate the predicted boxes. Different from RPN that predicts region proposals ($2$ classes) and classifies all proposals in another stage, our model predicts the final classes for all objects ($c$ classes). This change works well for biomedical images, while saving memory, because compared to natural scene images, there can be less classes but more instances in one stack (e.g., some types of cells). Thus, changing RPN to an object detector can help focus on locating instances and classifying instances into different classes in one step instead of two steps, which uses less parameters and less GPU memory.
We use the same multi-task loss of RPN for our detected boxes, i.e., $L_{box} = L_{cls} + L_{reg}$, but $L_{cls}$ is for all $c$ classes, not only two classes (objects and background). 
Since all instances have box annotation, the model conducts back-propagation for all boxes (Fig.~\ref{flow}), i.e., all boxes contribute to the object detector.


\subsection{3D voxel segmentation using full voxel  annotation for a small fraction of instances}

\noindent
\textbf{Mask-RCNN}. Mask-RCNN \cite{he2017mask} is used to perform 2D instance segmentation based on RPN. After generating RoIs using RPN, Mask-RCNN uses RoIAlign to align all RoIs to the same size (e.g., $7\times 7$). RoIAlign computes the value of each sampling point by bilinear interpolation from the nearby grid points on the feature maps. Using the interpolation value of the nearby grid points instead of using the value of one nearest point can make the predicted mask smoother and more accurate.
Then up-convolutional layers are used to calculate the mask for all aligned feature maps.

\noindent
\textbf{3D voxel segmentation}.
After detecting the instances for all objects, a 3D voxel segmentation model utilizing full voxel annotation for a small fraction of the instances is used to segment all instances. 
The corresponding feature maps are cropped from the feature layers of the backbone. To make the detected objects of different sizes share the same segmentation parameters, we extend the RoIAlign design from 2D to a 3D version. All cropped features are aligned to size $s \times s \times s \times p$, where $s$ is the RoIAlign size and $p$ is the channel number of the feature maps. For each sampling point, we first find the 8 nearest neighbor points on the feature maps, and then apply trilinear interpolation to the 8 neighbor points to calculate the value of the sampling point. Then up-convolutional layers are applied to all aligned features for segmenting all the detected instances. To make the model utilize the information from different layers, our 3D RoIAlign is applied to two different layers from the backbone, and the feature maps are concatenated as the final feature maps for 3D RoIAlign. (Fig.~\ref{backbone}). 

Since full annotation methods can be impractical for 3D instance segmentation, our model needs full voxel annotation only for a small fraction of the instances.
We add a mask-weight layer before calculating the loss for the voxel masks. Although the model segments all instances (no matter there is corresponding voxel annotation or not), it conducts back-propagation only for those instances having voxel mask annotation (Fig.~\ref{flow}).
We add a mask weight layer to set the loss to 0 for these instances without voxel annotation; by this means, only the instances with full voxel annotation contribute to the voxel segmentation model, and all other instances do not affect the voxel segmentation model. The loss for the mask $L_{mask}$ is the average binary cross-entropy loss. The loss for the whole task is $L = L_{box}+L_{mask}$. Experimental results show that, by adding the mask-weight layer, our 3D segmentation model can segment all detected instances utilizing full voxel annotation for only a small fraction of instances.


\section{Experiments and Results}

We evaluate our 3D instance segmentation model and our weak annotation method on three biomedical image datasets: nuclei of HL60 cells \cite{ulman2017objective}, microglia cells (in-house), and C.elegans developing embryos \cite{ulman2017objective}. For nuclei of HL60 cells and C.elegans developing embryos, our objective is different from the original challenge; we only use the data with ground-truth labels as both the training data and test data. We evaluate both the instance detection and instance segmentation performance on the nuclei of HL60 cells and microglia cells. Due to lack of full voxel annotation for the C.elegans embryo dataset, we only evaluate the instance detection performance on this dataset. Based on our experiments, it takes about 15GB GPU memory during training when a batch contains 4 stacks of size $64\times64\times128$ each. During testing, a stack of size $64\times639\times649$ containing 20 instances takes about 10 seconds on an NVIDIA Tesla P100 GPU.  


\noindent
\textbf{Nuclei of HL60 cells and microglia cells.}
Both these two datasets have full voxel annotation for all instances. For these two datasets, the time for an expert to label all the voxels of a cell is about 30 times of that for labeling a 3D bounding box of the cell according to our annotation time statistics. The dataset of HL60 cells has two groups of data: 150 stacks for the 1st group, and 80 stacks for the 2nd group. We use stacks $000, 010, 020, \ldots, 070$ from the 1st group and stacks $000, 010, 020, 030, 040$ from the 2nd group as the training data (13 stacks in total), and use stacks $080-149$ from the 1st group and stacks $050-079$ from the 2nd group as the test data (100 stacks in total).
For microglia cell images, we use 10 stacks of the 14 total stacks as training data, and the other 4 stacks as test data. For both the datasets, the number $c$ of classes is 2.

To evaluate our 3D voxel segmentation model utilizing full voxel annotation for a small fraction of the instances, for the HL60 cells images, we randomly choose $20\%$ of the cells from each stack as the instances with full voxel masks, and $30\%$ for microglia cells. All the instances have box annotation.

A best-known deep learning method is selected for comparison with our method. VoxResNet \cite{chen2016voxresnet} is applied to the training data with boundary class to produce voxel segmentation results, and then 3D connected components are computed from the voxel segmentation results as the instance segmentation results. 
For the comparison method, we evaluate the performance of using full voxel annotation and the performance of using similar annotation time as our method. 

For instance detection evaluation, we compute only whether the detected 3D bounding boxes match with some 3D ground-truth bounding boxes. All the detected boxes with IoU larger than 0.4 (note that 0.4 in 3D is more strict than 0.5 in 2D) with some matched ground-truth boxes are taken as true positives. All the other detected boxes are false positives. All ground-truth boxes without matching detected boxes are false negatives. For instance segmentation evaluation, we follow a similar evaluation process as in \cite{yang20163d}.


Table~\ref{tab:nuclei} and Table~\ref{tab:glia} show the results on these two datasets, in which the methods either use a proportion of the full voxel annotation (for our method, all instances have box annotation, and for the comparison method, full annotation is used) or use full annotation in the experiments.

\begin{table}[t]
    \setlength{\tabcolsep}{2pt}
    \renewcommand{\arraystretch}{0.9}
	\centering
	\caption{Results on the  
	HL60 cells dataset. AT = approximate annotation time.}
	\begin{tabular}{| c || c || c | c | c || c | c | c|}
	  \hline
	  \multirow{2}{*}{Method} & \multirow{2}{*}{AT} &\multicolumn{3}{c||}{Detection F1} & \multicolumn{3}{c|}{Segmentation F1} \\
	  \cline{3-8}
	  & & Group 1 & Group 2 & Mean & Group 1 & Group 2 & Mean \\
	  \hline
	  Our method ($20\%$)&5.5h & 0.9967 & \textbf{0.9599} & \textbf{0.9783} & 0.9416 & \textbf{0.8437} & 0.8927\\ \hline
	  
	  VoxResNet (full) & 22.5h & \textbf{0.9988}  & 0.9543 & 0.9766 & \textbf{0.9656}  & 0.8428 & \textbf{0.9042} \\ \hline
	  
	  VoxResNet ($4/13$) &8.3h & 0.9965 & 0.9221 & 0.9593 & 0.9610 & 0.7873 & 0.8742\\ \hline
	  
	\end{tabular}
	\label{tab:nuclei}
\end{table}

\begin{table}[t]
    \setlength{\tabcolsep}{4pt}
    \renewcommand{\arraystretch}{0.9}
\centering
\small
\caption{Results on the microglia cells dataset. AT = approximate annotation time.}\label{tab:glia}
\begin{tabular}{|c||c||c||c|} \hline
                         Method & AT & Detection F1 & Segmentation F1 \\ \hline
Our method ($30\%$) & 54.7h & \textbf{0.9078} & 0.8424 \\ \hline

VoxResNet (full)  & 172.9h &  0.9017 & {\textbf{0.8484}} \\ \hline
VoxResNet ($4/10$)  & 67.5h & 0.8761 & 0.8307\\ \hline

\end{tabular}
\end{table}

For instance detection, our method using full voxel annotation for only a small part of the instances outperforms the comparison method using full annotation. This is because box annotation contains stronger instance information than voxel annotation, and our model considers the loss for instance detection explicitly ($L_{box}$). For instance segmentation, due to the resampling operations in RoIAlign, the comparison method using full annotation is better than ours, but ours still outperforms the comparison method with similar annotation time, because locating instances accurately can improve the performance of voxel segmentation. Although our method dose not surpass the comparison method using full annotation, it is still practical due to using much less annotation time.

\noindent
\textbf{C.elegans developing embryos.}
For this dataset, we do not have full voxel annotation but only small markers indicating different instances. The dataset is quite dense (e.g., see Fig.~\ref{fig:gt_test}), and it is difficult (or impractical) to label all voxels of all instances. This dataset has two groups of data; each group contains 190 stacks with markers. To evaluate our detection method, experts labeled 1527 3D bounding boxes for 15 stacks from the first group, and only 67 instances are labeled with full voxel annotation. We use the 190 stacks in the second group for testing.
A sample result is given in Fig.~\ref{fig:gt_test}(c)-(d). We determine the performance of our instance detection by computing the distances between the ground-truth markers and the centers of the detected boxes. The F1 score for this experiment is $0.9495$ (using a distance threshold of 5 pixels).


\section{Conclusions}
In this paper, we presented a new end-to-end 3D instance segmentation approach, and to reduce annotation effort on 3D biomedical images, we proposed a weak annotation method for training our 3D instance segmentation model. Experimental results show that (1) with full annotated boxes and a small amount of masks, our approach can achieve similar performance as the best-known methods using full annotation, and (2) with similar annotation time, our approach outperforms the best-known methods that use full annotation.

\subsubsection{Acknowledgment.}
This research was supported in part by NSF grant CCF-1640081 and the Nanoelectronics Research Corporation (NERC), a wholly-owned subsidiary of the Semiconductor Research Corporation (SRC), through Extremely Energy Efficient Collective Electronics (EXCEL), an SRC-NRI Nanoelectronics Research Initiative under Research Task ID 2698.005, NSF grants CCF-1617735 and CNS-1629914, and NIH grant R01 R01CA194697.

\bibliographystyle{splncs03}
\bibliography{paper1252}
\end{document}